\documentclass[]{fairmeta}

\usepackage{outlines}
\usepackage{amssymb}
\usepackage{pifont}
\usepackage{adjustbox}
\usepackage{caption}
\usepackage{graphicx}
\usepackage{array}
\usepackage{multirow}
\usepackage{graphicx}
\usepackage{scrextend}

\definecolor{Red}{rgb}{1,0,0}
\definecolor{Green}{rgb}{0,0.69,0}
\definecolor{Blue}{rgb}{0,0,1}
\definecolor{LightBlue}{rgb}{0,0.5,1}
\definecolor{veryLightBlue}{rgb}{0.85,0.98,1}
\definecolor{veryLightGreen}{rgb}{0.6,1,0.6}
\definecolor{Skin}{rgb}{1,0.71,0.69}
\definecolor{Grey}{rgb}{0.5,0.5,0.5}
\definecolor{LightGrey}{rgb}{0.6,0.6,0.6}
\definecolor{VeryLightGrey}{RGB}{219, 219, 219}
\definecolor{Black}{rgb}{0,0,0}
\definecolor{White}{rgb}{1,1,1}
\definecolor{brickred}{rgb}{0.8, 0.25, 0.33}
\definecolor{burntOrange}{RGB}{255,122,20}
\definecolor{navy}{RGB}{80, 74, 255}
\definecolor{teal}{RGB}{0, 123, 159}
\definecolor{aquamarine}{RGB}{51, 153, 255}
\definecolor{saffron}{RGB}{227, 170, 0}
\definecolor{purplePink}{RGB}{160, 89, 107}
\definecolor{xanadu}{RGB}{126, 145, 129}
\definecolor{deepred}{RGB}{217, 18, 4}
\newcommand{\red}{\color{Red}}

\newcommand{\burntOrange}{\color{burntOrange}}
\newcommand{\lightBlue}{\color{LightBlue}}

\newcommand{\deepRed}{\color{deepred}}

\definecolor{realImageColor}{RGB}{8, 194, 150}
\definecolor{synImageColor}{RGB}{24, 166, 237}
\newcommand{\realImageColor}{\color{realImageColor}}
\newcommand{\synImageColor}{\color{synImageColor}}

\newcommand{\eg}{\emph{e.g.}\ }

\newcommand{\dataset}{\texttt{Common-O Bench}}
\newcommand{\datasetComplex}{\texttt{Common-O Complex}}
\newcommand{\Cross}{\textcolor{red}{\times}}
\newcommand{\CheckTrue}{\textcolor{green}{\checkmark}}

\title{What’s in Common? Multimodal Models Hallucinate When Reasoning Across Scenes}

\author{Candace Ross}
\author{Florian Bordes}
\author{Adina Williams}
\author{Polina Kirichenko}
\author{Mark Ibrahim}

\affiliation{\textbf{FAIR at Meta}}

\abstract{
Multimodal language models possess a remarkable ability to handle an open-vocabulary's worth of objects. Yet the best models still suffer from hallucinations when reasoning about scenes in the real world, revealing a gap between their seemingly strong performance on existing perception benchmarks that are saturating and their reasoning in the real world. To address this gap, we build a novel benchmark of in-the-wild scenes that we call \dataset. With more than 10.5k examples using exclusively new images not found in web training data to avoid contamination, \dataset\ goes beyond just perception, inspired by cognitive tests for humans, to probe reasoning across scenes by asking ``what’s in common?''. We evaluate leading multimodal language models, including models specifically trained to perform chain-of-thought reasoning. We find that perceiving objects in single images is tractable for most models, yet reasoning across scenes is very challenging even for the best models, including reasoning models. Despite saturating many leaderboards focusing on perception, the best performing model only achieves 35\% on \dataset---and on \datasetComplex, consisting of more complex scenes, the best model achieves only 1\%. Curiously, we find models are more prone to hallucinate when similar objects are present in the scene, suggesting models may be relying on object co-occurrence seen during training. Among the models we evaluated, we found scale can provide modest improvements while models explicitly trained with multi-image inputs show bigger improvements, suggesting scaled multi-image training may offer promise. We make our benchmark publicly available to spur research into the challenge of hallucination when reasoning across scenes.
}

\date{\today}
\metadata[HuggingFace Dataset]{\url{https://huggingface.co/datasets/facebook/Common-O}}

\begin{document}

\maketitle

\section{Introduction}

\begin{figure}
    \centering
    \includegraphics[width=1.08\linewidth]{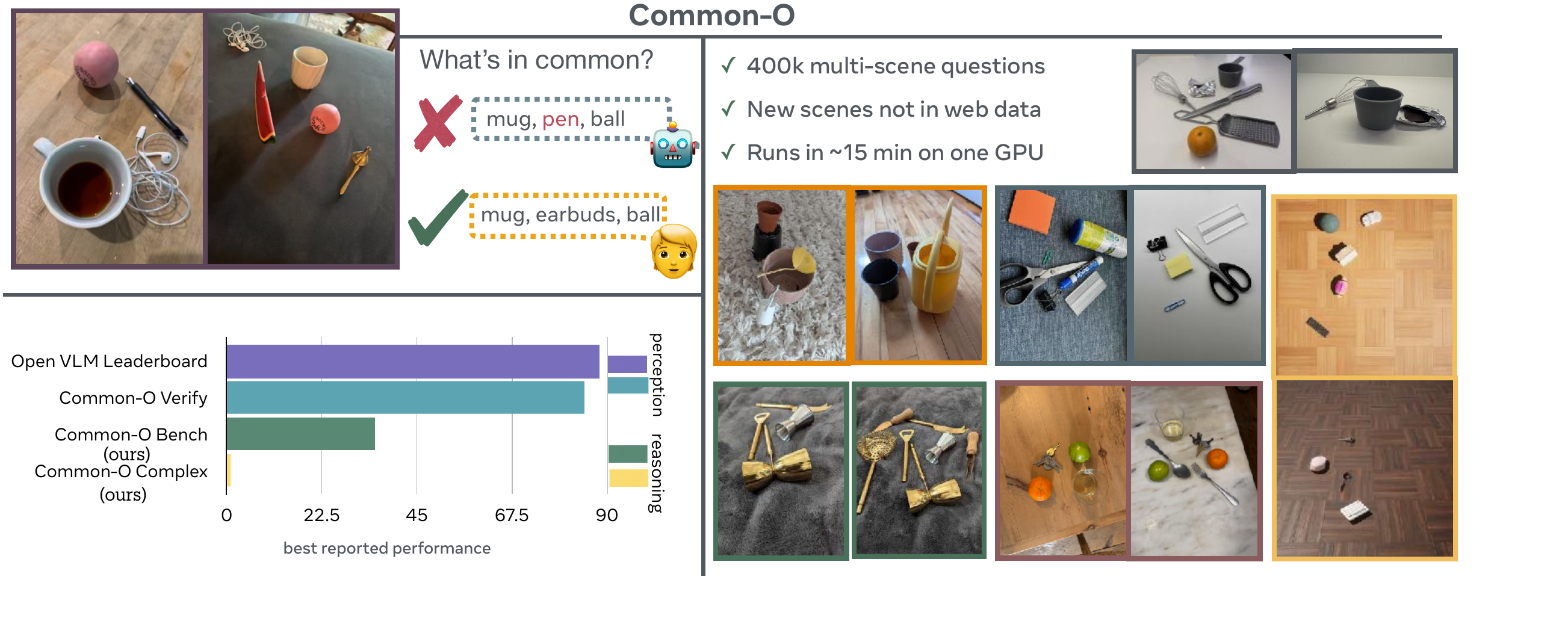}
    \caption{\textbf{Reasoning across scenes is an open challenge for today's best multimodal models.} We show the best performance from the Open VLM leaderboard on MMBench and single image evaluations from our benchmark illustrating saturation for perception tasks.}
    \label{fig:enter-label}
\end{figure}

Multimodal models today are starting to saturate visual perception leaderboards. 
For example, on classical text-and-image benchmarks such as CLEVR \citep{johnson_clevr_2016}, DocVQA \citep{mathew-etal-2021-docvqa}, ChartQA \citep{masry-etal-2022-chartqa}, TextVQA \citep{singh-etal-2019-towards}, MMBench \citep{liu2023mmbench}, and Seed-Bench \citep{li2024seed}, top-performing models achieve an accuracy of 80\%-90\% \citep{zhang2024mme0realworld0}. However, despite these impressive results, there is a growing concern that these benchmarks may not accurately reflect the performance of models in real-world settings. In fact, research has shown that models often struggle to generalize to new, unseen data, and are prone to hallucinating objects that are not present in the scene \citep{guan2023hallusionbench0}.

One of the main reasons for this gap between benchmark performance and real-world performance is the contamination between benchmarks and training data \citep{chen2024we}. Many benchmarks are constructed from web data, which is also used to train models. This means that models are often trained on data that is quite similar to the benchmark data, leading to an overestimate of their performance. Another factor contributing to the gap is the lack of benchmarks that capture the complexity and variability of real-world scenes. Many benchmarks use simplified geometric visuals or static scenes; while useful for static tasks such as chart understanding, these do not reflect the dynamic and evolving nature of real-world scenes.

The ability to reason across complex scenes containing multiple objects is a fundamental aspect of human cognition. Research in cognitive science has shown object identification in scenes is a key component of cognitive function---and that deficits in this ability are a hallmark of cognitive decline, such as in Alzheimer's disease \citep{SceneryPicture}. Furthermore, studies have demonstrated that the brain's ability to understand the relationships between objects in a scene is closely tied to its ability to understand the scene as a whole \citep{damasio-etal-2001-neural, brandman-etal-2017-interaction}. These findings suggest that a benchmark that requires models to reason about complex, dynamic scenes is crucial step towards deploying reliable models in the real world.

To address these challenges, we introduce a new multi-image benchmark,~\dataset, designed to test models' ability to reason \textit{across dynamic scenes} in a way more similar to human reasoning. Our benchmark includes scenes containing multiple objects with varying lighting conditions, and complex backgrounds that requires models to reason about the relationships between objects across distinct scenes. We choose scenes with up to 7 objects as our default setting, inspired by the 1956 classic, putative, heuristic constraint on human memory, colloquially described as the ``Magical Number Seven, Plus or Minus Two" (\citealt{magicalnumber, baddeley-1994-magical, cowan-etal-2007}, i.a.).  \dataset\ comprises both real and synthetic data, allowing for more flexibility in our evaluation, as we can sample a wide range of object-background combinations that are typical in real data. We also provide a non-overlapping fully synthetic challenge set, \datasetComplex\ that spans up to 16 objects per scene, increasing scene complexity appreciably. In both \dataset\ and \datasetComplex, we intentionally provide multiple camera points of view of a given scene, reflecting the diversity found in the real world.

We find that despite being able to recognize objects in individual scenes, state-of-the-art models struggle to reason across scenes. The best performing model we tested, GPT-4o, achieves only 35\% on \texttt{Common-O Bench}, highlighting reasoning across scenes as open challenge, in stark contrast to the saturation observed for other multimodal benchmarks. For the more challenging set \texttt{Common-O Complex}, the best performing model achieve $<$1\%. Curiously, we find models hallucination is \textit{pervasive}, with at least one 1 object hallucinated 53\% of the time and 2+ as often as 23\% of the time.  

Our findings have important implications for the development of multimodal models~\citep{bordes2024introductionvisionlanguagemodeling}. We find models trained with multi-image inputs achieve higher performance and scale can yield marginal benefits, yet even the best multi-image large scale models struggle highlighting the need for models to be designed with real-world scenes in mind. This requires a fundamental shift in the way models are designed and trained, and underscores the need for more research in this area. We release \texttt{Common-O Bench} and \texttt{Common-O Complex} to mark a new challenge in multimodal models' ability to reason across scenes that we hope could unlock new frontiers in real world applications \footnote{Datasets are available at \url{ https://huggingface.co/datasets/facebook/Common-O}}.

\section{Related Work}

Many works have aimed to evaluate model performance on visual reasoning. We summarize our contributions relative to existing benchmarks in terms of multi-image capability, scale, and saturation in \Cref{tab:benchmark-comparison}. Our dataset is larger in size, captures multi-image reasoning across scenes inspired by human cognitive tests, and stands out in terms of not relying on existing web datasets, thereby avoiding possible training data contamination or object resemblance. Together, these factors make our benchmark much more challenging relative to existing benchmarks where performance has saturated. 

\begin{table}[!ht]
\small
\centering
\begin{tabular}{llccrcr}
\toprule
& \textbf{Benchmark} & \textbf{Multi-image} & \textbf{Multi-scene} & \textbf{Size} & \textbf{Source} & \textbf{SOTA} \\
\midrule
\multirow{4}{*}{\rotatebox{90}{\scriptsize Abstract}} & NTSEBENCH & $\Cross$ & $\Cross$ & 2.7k & Web & 88.9\% \\
& MathVista & $\Cross$ & $\Cross$ & 6k & Existing \& new & 80.9\% \\
& MMIU Objective Semantic& $\CheckTrue$ & $\CheckTrue$ & 1.2k & Existing & 55.7\% \\
& ReMI & $\CheckTrue$ & $\CheckTrue$ & 2.6k & Synthetic & 50.5\% \\
\midrule
\multirow{2}{*}{\rotatebox{90}{\scriptsize Hallu.}}
& POPE & $\Cross$ & $\Cross$ & 9k & Existing 
& 91.0\% \\
& HallusionBench & $\CheckTrue$ & $\CheckTrue$ & 591 & Synthetic \& cartoon & 67.1\% \\
\midrule
\multirow{5}{*}{\rotatebox{90}{\scriptsize Real}}
& MMBench & $\Cross$ & $\Cross$ & 1784 & Web & 88.3\% \\
& NLVR2 & $\CheckTrue$ & $\CheckTrue$ & 13.9k & Web & 80.3\% \\
& GQA & $\CheckTrue$ & $\CheckTrue$ & 3.4k & Existing 
& 74.6\% \\
& SEED-Bench-2 & $\CheckTrue$ & $\CheckTrue$ & 660 & Existing 
& 73.1\% \\
& MUIRBench & $\CheckTrue$ & $\CheckTrue$ & 536 & Existing \& new 
& 68\% \\
\midrule
\multirow{2}{*}{\rotatebox{90}{\scriptsize \textbf{Ours}}}
& \texttt{Common-O Bench} & $\CheckTrue$ & $\CheckTrue$ & 10k & New & 35\% \\
& \texttt{Common-O Complex} & $\CheckTrue$ & $\CheckTrue$ & 12k & New & 1\% \\
\bottomrule
\end{tabular}
\caption{Existing benchmark datasets targeting abstract reasoning, hallucination (`Hallu.'), and real image reasoning) are insufficient due to saturation, and/or failure to target multi-image and/or multi-scene reasoning. Existing datasets targeting multi-image and multi-scene reasoning exist but have saturated (NLVR2, GQA). Those that have not saturated are relatively small (SEED-Bench-2, MUIRBench, HallusionBench, ReMI). Abstract benchmarks mostly focus on abstract geometric reasoning in puzzles/charts rather than real scenes or extract frames from videos.}
\label{tab:benchmark-comparison}
\end{table}

\paragraph{Perception.}
Many benchmarks include composite measures that focus on single object-centric perception such as object classification \citep{deng2009imagenet,lin2014microsoft} and attributes or relations of objects \citep{al-tahan_unibench_2024, dumpala_sugarcrepe_2024}. 
As part of perception, researchers have also focused on the contribution of the background to object identification \citep{beery-etal-2018-recognition,sureddy-etal-2024-decomposed}, as well as issue of hallucination where models describe objects that are not present in scenes \citep{li2023evaluating,guan2023hallusionbench0}. Instruction following \citep{li2023mimicitmultimodalincontextinstruction} for perception tasks using single images is another area where diversity, quality, and creativity of answers is important.
To assess the robustness of perception capabilities, researchers have also used synthetic generation to vary object attributes and compose diverse scenes \citep{bordes_pug_2023,gan2021threedworldplatforminteractivemultimodal}.
Recent efforts to benchmark multimodal model have relied on larger composite suites of benchmarks that span several tasks such as  recognition, OCR, counting, visual question answering, and  object attributes etc. \citep{yu2023mmvet,liu2023mmbench,li2024seed}.

\paragraph{Abstract reasoning in charts, geometric sketches, and puzzles.}
Relative to the improved performance on real world perception tasks, multimodal models exhibit degraded performance on abstract visual puzzles that involve straight-forward reasoning. 
For example, \citet{cho_vision_2025} show multimodal models lag considerably behind humans at identifying simple tasks such as whether two circles overlap, with \citet{huang2025vision} showing similar conclusions on visual arithmetic.
Similarly, \citet{wust_bongard_2025, jiang_marvel_nodate,ullman_illusion-illusion_2024,kraaijveld_columbus_2024} probe whether models can solve basic visual logical puzzles that involve outlines of geometric shapes, illusions, and lateral thinking.
\citet{pandya_ntsebench_2025} construct a dataset of 2.7k multiple choice questions from the national exam in India that involve geometry and visual reasoning questions from graphs. \citet{hemmat_hidden_nodate} evaluates whether multimodal models can perceive abstract shapes, a key aspect of human visual perception. \citet{sampat_vl-glue_2024} assess whether multimodal models can solve NLP and visual tasks jointly. 
\citet{lin2024comparisonvisualinstructiontuning} studies comparisons across pairs of synthetically generated CAD images.
Most similar to our work is the objective high-level semantic task from MMIU, which consist of 1.1k examples from existing datasets focused on semantic correspondence such as BLINK \citep{fu2024blink0} and MISC210K \citep{sun2023misc210k}, spotting the difference \citep{jhamtani2018learning} or abstract puzzles from datasets such as NLVR2 \citep{suhr-etal-2019-corpus}. We build on this setup to focus on reasoning about object commonality across scenes at larger scales.

\paragraph{Measuring reasoning using single image benchmarks.} The prior benchmarks reveal abstract reasoning may be a challenge for multimodal models hinting at a possible reason for the observed gap in real world performance multimodal models. Many works attempted to measure the gap between real world capabilities and benchmark performance by focusing on robustness \citep{geirhos_imagenet-trained_2022,gabbay_image_2021,hendrycks_many_2021}. For example, \citet{richards_does_2023} measures the in-the-wild robustness gap for household object classification across geographies. Another approach to capture the real world versus standard benchmark gap is to explicitly mine or generate challenging images \citep{tong_mass-producing_2024,tong_eyes_2024,wang_journeybench_2025}. A first step to reasoning beyond perception is compositionality. Several works have studied whether multimodal models can understand and compose attributes and objects \citep{johnson_clevr_2016,thrush_winoground_2022,yuksekgonul2022and,krojer2022image,kil_mllm-compbench_2025, wu2025compact0}. Some have even explored single-image reasoning in adversarial settings \citep{li-etal-2021-adversarial, sheng-etal-2021-human} and memes \citep{kiela2020hateful,suryawanshi2021findings}. Yet, real world scene understanding requires reasoning beyond basic single-image settings.

\paragraph{Multi-image reasoning across natural scenes.} To go beyond perception, \textit{generalization in the real world requires reasoning across scenes}. 
Aggregate benchmark such as VHELM contain multi-image reasoning tasks \citep{lee_vhelm_2024}, many of which are derived from geometry style puzzles akin to those described above. The real world image reasoning task in VHELM is based on GQA, which is a dataset constructed from objects in the popular Visual Genome dataset available in web training data \citep{agrawal-etal-2022-alpha}. The authors use 11k images from the GQA validation set in their evaluations. There are also multi-image binary tasks with image selection \citep{hu2019evaluating}, predicting whether captions are true of images \citep{suhr2017corpus,suhr-etal-2019-corpus}, and  visual haystacks \citep{wu2024visual} that focuses on retrieval as well as visual question answering based on a large number of images (up to 10k). Other benchmarks \citep{meng2024mmiu0,fu2024blink0} also focus on multi-image tasks, showing even models that excel at single image tasks struggle on multi-image tasks such as visual correspondence, semantic correspondence, and multi-view reasoning of the same scene across multiple images. Other tasks include visual similarity, relative depth, and functional correspondence in the same image. However, as shown in \Cref{tab:benchmark-comparison}, these multi-image real scene benchmarks rely on mining image from the web or existing datasets, which both limits their size and introduces possible training data contamination. We observe the best reported performance even on multi-image benchmarks is quite high 68-88.3\%.

\section{Methods}

\begin{figure}
\centering
\begin{subfigure}{\textwidth}
    \centering
    {What's in common?}\vspace{1ex}
\includegraphics[width=1.0\linewidth]{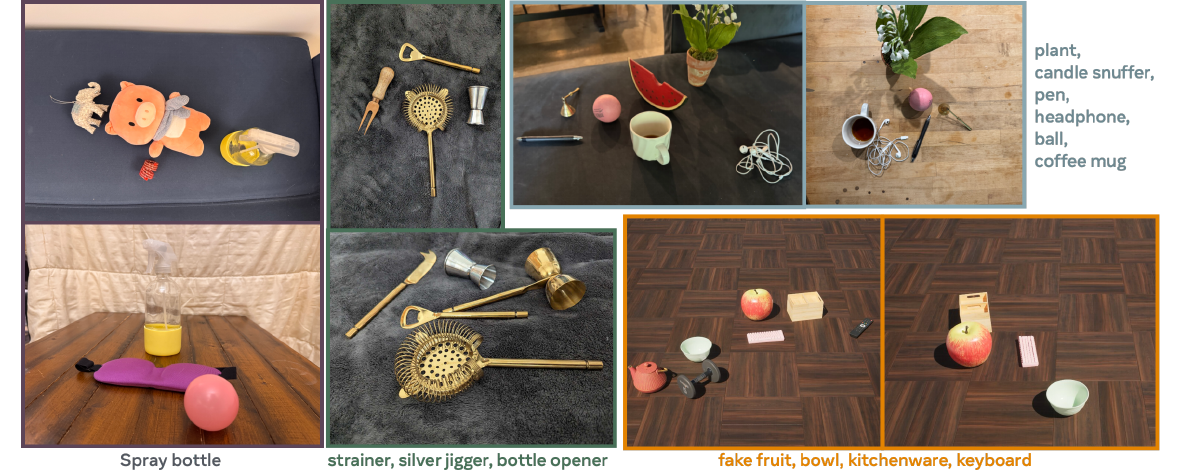}
\caption{\dataset: 10k examples of real and synthetic images, with scene complexity from 3 to 7 objects.}
\label{fig:main-dataset-example}
\end{subfigure}

\begin{subfigure}{\textwidth}
    \centering
    \includegraphics[width=1.0\linewidth]{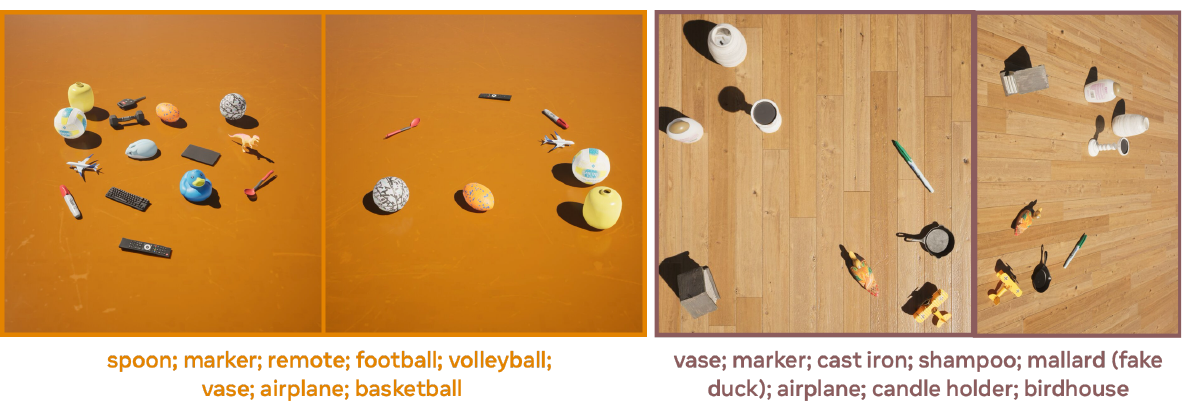}
    \caption{\datasetComplex: 12k containing synthetic images only, ranging in complexity from 8 to 16 objects.}
    \label{fig:challenge-set-example}
\end{subfigure}
\caption{\dataset\ contains real and synthetic images of objects in different orientations and configurations. These are randomly selected examples from the dataset along with the human ground truth labels for the common object(s) between them. }

\label{fig:all-dataset-examples}
\end{figure}

\subsection{Dataset Construction}\label{sec:dataset}

\dataset\ is designed to test the ability of models to reason about complex, dynamic scenes in a way that is similar to human reasoning. \dataset\ consists of 10.5k examples, representing different scenes containing 3 or more objects with diverse background and viewpoints. 
configurations of objects. 
Every example in the dataset consists two images, which can be either real (45\%) or synthetic (55\%). To ensure that each image is completely new and unique, with no issues of contamination in web or existing data used for training, the real images were taken by four experts in machine learning with no particular photography training. Image-takers followed a fairly simple data creation procedure where images were grouped in sets and placed arbitrarily against simple backgrounds to generate test data. We do not include any images of people or proprietary content such as brands or logos. See \Cref{app:takingpictures} for more details on our image-taking guidelines. The synthetic images were generated using Unreal Engine 5.4 with assets from the Aria Digital Twin Catalog \citep{Dong_2025_CVPR}. We place the objects randomly in the scene and take pictures from different angles. To avoid any overlapping of objects, we rescale each of image to a given maximum size while maintaining their aspect ratio (more detail can be found in \Cref{app:syntheticdata}).
We also construct \datasetComplex\ consisting of 12k examples of more complex scenes, containing $8 - 16$ objects, and being wholly synthetically created using the same video game engine. This allows us to evaluate the ability of models to reason about scenes with varying levels of complexity and artificiality. 
We have 129 different objects in \dataset\ and \datasetComplex. Using Segment Anything \citep{kirillov2023segany}, we find the object in the images ranges from 2-22\% of the overall image size. Following \citet{gebru2021datasheets}, we include a full dataset card in \Cref{sec:dataset-card}.
See \Cref{fig:all-dataset-examples} for examples from \dataset\ and \datasetComplex.

\subsection{Evaluation}

\paragraph{Task Definition.} An input example is defined as $(I_0, I_1, \mathcal{O}_{\text{choices}}, \mathcal{O}_{\text{in\_common}})$ where:
\begin{itemize}
    \item $I_0, I_1$ are the two images $I_0, I_1$
    \item $\mathcal{O}_{\text{choices}}$ is a set of candidate objects $\mathcal{O}_{\text{choices}}$
    \item $\mathcal{O}_{\text{in\_common}}$ is the set of ground truth objects in common between the images.
\end{itemize}

Models are tasked with predicting the common objects $\mathcal{O}_{\text{in\_common}}$. We format the data $(I_0, I_1, \mathcal{O}_{\text{choices}})$ into model input.

To isolate perception from reasoning capabilities, we conduct a single-image evaluation as well. Models receive one image and a binary question (\textit{``Is <object> in this image?''}), testing basic object recognition. Strong performance here suggests failures in multi-image setups stem from reasoning limitations rather than perception deficits. This controlled comparison enables clearer analysis of cross-image reasoning abilities. We also performed human annotations with 4 expert annotators who are authors using 100 randomly sampled examples (each reviewed by at least two annotators). We reach 84\% human annotation agreement. 

\paragraph{Metrics.} We assess performance through two complementary metrics. First, \textbf{accuracy} measures strict correctness, requiring an exact match between predicted ($\mathcal{O}_{\text{pred}}$) and ground truth ($\mathcal{O}_{\text{common}}$) object sets. Second, \textbf{hallucination rate} quantifies how often model respond with an object that is not present. Specifically, hallucination measures the false positive predictions, calculated as the ratio of incorrectly predicted objects to total choices: $\frac{|\mathcal{O}_{\text{pred}} \setminus \mathcal{O}_{\text{common}}|}{|\mathcal{O}_{\text{choices}}|}$. This combination enables evaluation of both precision and recall in model predictions. 

\paragraph{Models.} We benchmark a diverse array of multimodal models spanning different architectural families and scales. Openly available models include LLaVA-OneVision (7B, 72B) \citep{li2024llavaonevision}, DeepSeek-VL2 (Small/Base) \citep{wu2024deepseekv2}, LlamaV-o1 \citep{thawakar2025llamav}, Qwen2.5-VL \citep{Qwen2.5-VL}, LLaMA-4 Scout Instruct \citep{llama4}, PerceptionLM (3B/8B) \citep{cho_perceptionlm_2025} and QVQ-72B-Preview \citep{qvq-72b-preview}. The closed-source GPT-4o is also evaluated\footnote{Note that we use a slightly different prompt setup for GPT-4o, where the model predicts object values instead of letters. We provide the full comparison in the appendix.}. Our implementation uses HuggingFace Transformers \citep{wolf2020transformers} for LLaMA-V-o1, the Perception Models GitHub repository\footnote{\url{https://github.com/facebookresearch/perception_models}} for PerceptionLM, and vLLM \citep{kwon2023efficient} for remaining models. We ran all models locally, on single node with 8 A100s GPUs, except for GPT-4o, which is only available through the API. All use greedy decoding with default parameters (temperature=1, top-p=1) unless specified otherwise. Images are resized, maintaining the aspect ratio, with the smallest size of $384$px. For models not explicitly trained for multi-image input---Llama 3.2, LlamaV-o1, PerceptionLM---we first concatenate the two images before passing them to the model as input.\footnote{For the best performing open source model, we additionally tested different temperatures and did not observe a significant performance difference. Results are shown in \Cref{app:temperature}.}

\paragraph{Model Input.} The object choices are alphabetized (A, B, C\textellipsis) to leverage models' preference for letter-based responses over other input formats \citep{long2024llms}. Outputs must conclude with a comma-separated prediction list, allowing flexible generation formats, including chain-of-thought reasoning \citep{wei-etal-2022-chain}. For models trained for multi-image input, text prompt is:

\begin{addmargin}[2em]{2em}
\texttt{Which objects are present in both images? Select all choices that are true: \{\}. You can think of your answer in any way (e.g. step-by-step) but for the last line of your response, respond only in this format `Answer: <letter 1> <letter 2> <letter 3>', e.g. `Answer: A, B, C'.
}
\end{addmargin}

For models where we first concatenate the input images, the text prompt is:
\begin{addmargin}[2em]{2em}
\texttt{There are two images provided, one on the left and the other on the right. Which objects are present in both images? Select all choices that are true: \{\}. You can think of your answer in any way (e.g. step-by-step) but for the last line of your response, respond only in this format `Answer: <letter 1> <letter 2> <letter 3>', e.g. `Answer: A, B, C'.
}
\end{addmargin}

We also tested two additional input prompt formulations, shown in \Cref{app:temperature}. We did not observe a meaningful performance difference across prompts.

\section{Results}

\subsection{Multimodal models can perceive, but struggle to discern what's in common across scenes.}

To evaluate the performance of various state-of-the-art models on \dataset, we first validate the difficulty of perception using a single image scene setup as shown in \Cref{fig:single-vs-multi-barplot}. 
We find all models exhibit strong performance on single-image perception, yet struggle to reason across the same scenes in \dataset. The best performing model, GPT-4o, achieves only 35\% accuracy with reasoning models performing even worse, highlighting the challenge of reasoning across scenes. Underlying many incorrect answer is a tendency for models to hallucinate objects. We also report standard errors, which we find to < 0.02\%, for all models across both single image perception and multi-image reasoning in \Cref{app:standard_errors}.

\subsection{Models often hallucinate objects when reasoning across scenes.}

\begin{figure}[h!]
\begin{subfigure}{\textwidth}
    \centering
    \includegraphics[width=\textwidth]{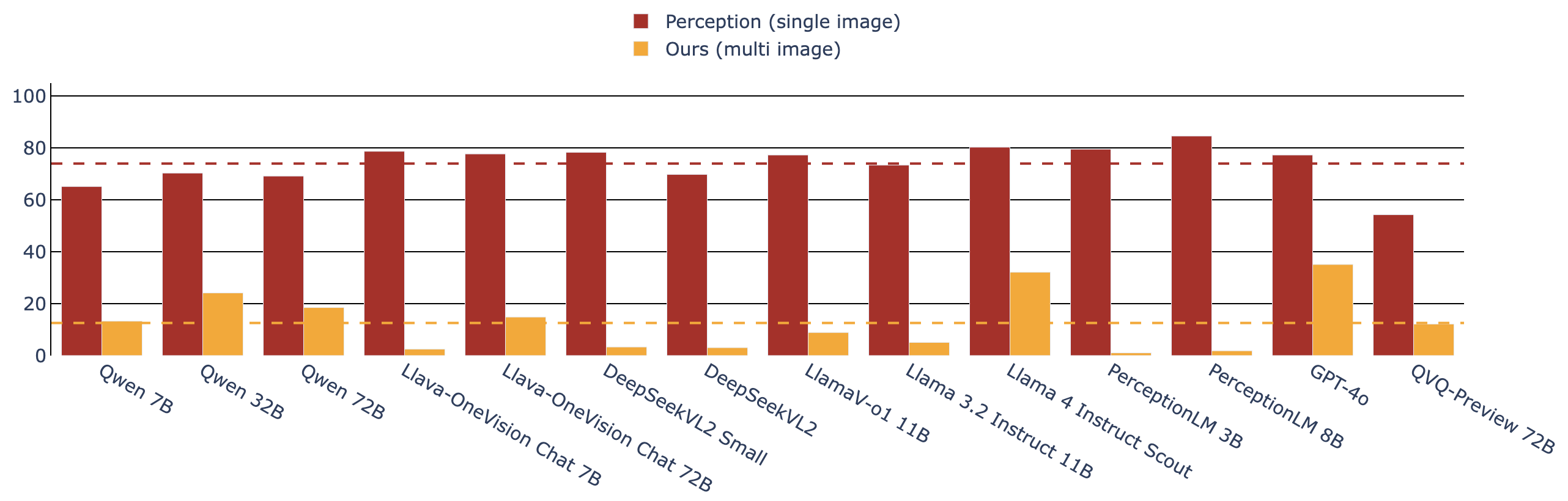}
    \caption{\textit{Accuracy ($\uparrow$):} Models can perform well in perceiving objects in the single image setting, but struggle to reason across scenes in our multi-image setting. The dashed lines show  performance averaged across models.}
    \label{fig:single-vs-multi-barplot}
\end{subfigure}

\begin{subfigure}{\textwidth}
    \centering\vspace*{1ex}
    \includegraphics[width=1\linewidth]{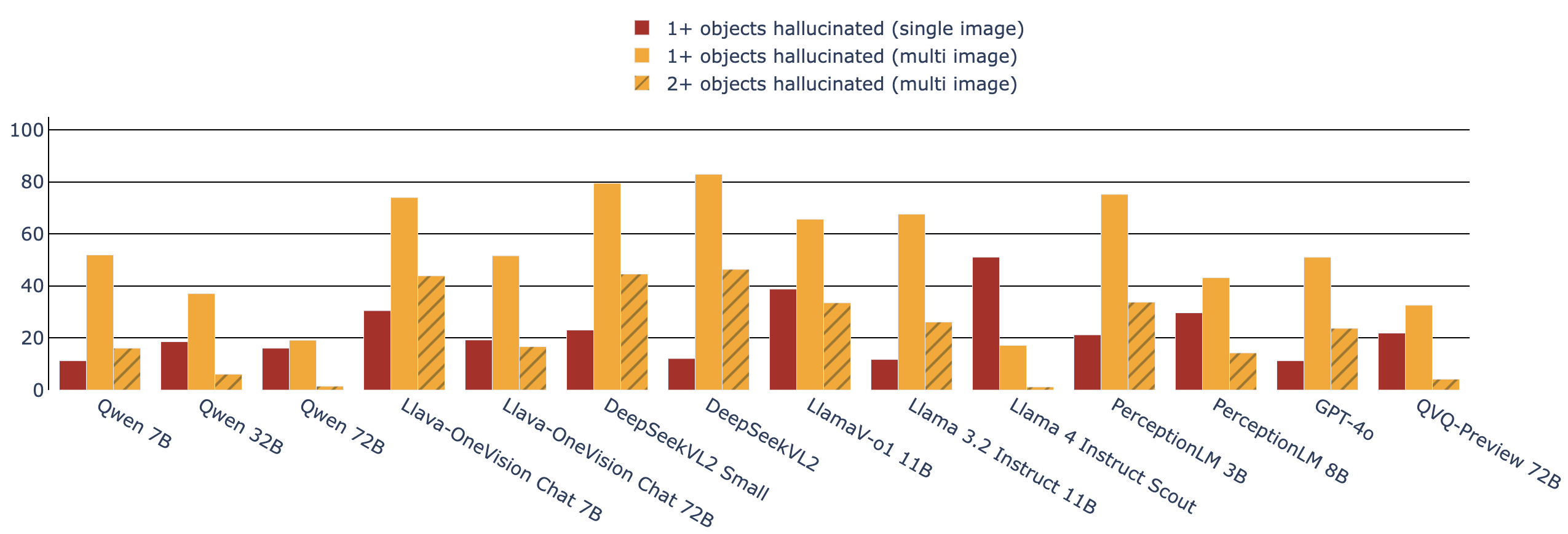}
    \caption{\textit{Rates of hallucination ($\downarrow$):} Models often hallucinate objects that are not present in the image. Models by and large hallucinate objects significantly more often in the multi-image case; Llama 4 was the only exception. Single image multiple object hallucination rates are not plotted as they were negligible for all models.}
    \label{fig:hallucination}
\end{subfigure}

\caption{Performance for single image object perception {\bf\burntOrange in yellow} and multi-image reasoning {\bf\deepRed in red} for (a) accuracy and (b) hallucination rates. We observe models with higher accuracy tend to also have lower rates of hallucination. We include a table of these results, along with statistical analyses, in \Cref{sec:stat-analysis-results}.}
\end{figure}

\begin{figure}
    \centering
        \includegraphics[width=1.0\textwidth]{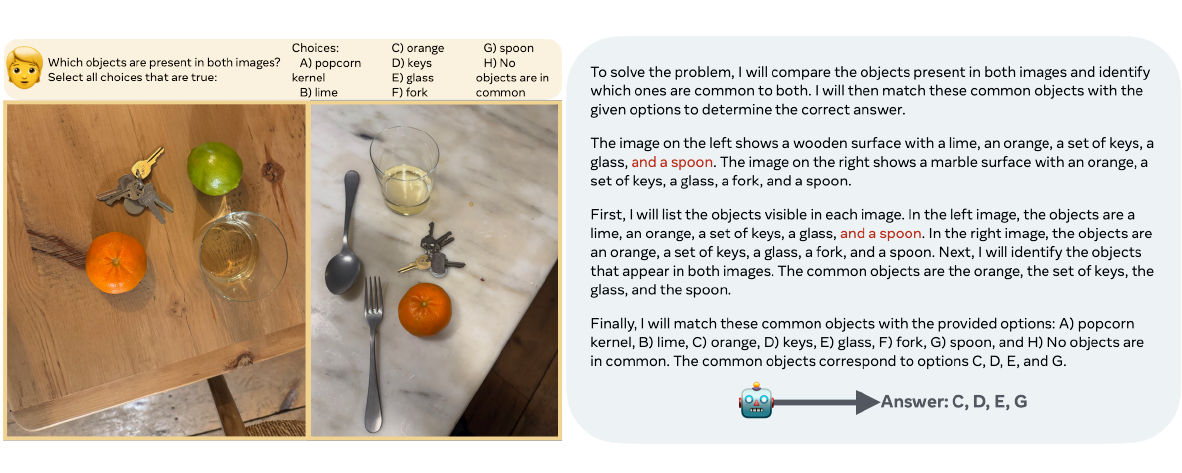}
    \caption{These are two examples of model failures, with the specific failures shown {\bf\red in red}.}
    \label{fig:failure-example}
\end{figure}

Models are very likely to hallucinate objects that are not present, which impacts their multi-scene reasoning abilities.
One possibility could be that, because of models' \textit{yes-}bias, they're likely to hallucinate objects that are not present when asked about them directly \citep{zhang2016yin,agrawal2018don,ross-etal-2024-what}. While models do hallucinate in this single image setting, \Cref{fig:hallucination} shows that hallucinate rates are significantly higher during the multi-image reasoning setting (excluding Llama 4, which does not follow this trend). 
We show an example hallucination in \Cref{fig:failure-example}. 
Similarly, hallucinating multiple objects rarely occurs in single image perception, but occurs more often when reasoning across scenes.

\paragraph{Model performance \textit{severely} degrades for more complex scenes.}

We use the challenge set, \datasetComplex, to explore how models perform with more complex scenes. \datasetComplex\ follows the same design described in \Cref{sec:dataset}, we just extended the scene complexity to $N$=8 to $N$=16 objects (see examples in \Cref{fig:challenge-set-example}). Across every model, performance severely drops. None of the models that we evaluated gets above $\sim$1\% accuracy.  We also see very high hallucination rates, with $76\%$ for 1+ objects and $55\%$ for 2+ objects on average across all models. This bolsters our decision to use a default setting of 7 objects as a good primary focus for models.

\begin{table}[h]
\centering
\begin{tabular}{c|ccccccccccccc}
    \toprule
    
    & \multicolumn{3}{c}{\small \underline{Qwen}}
    &
    \multicolumn{2}{c}{\small \underline{Llava Chat}}
    &
    \multicolumn{2}{c}{\small \underline{DeepSeek}}
    &
    \multicolumn{3}{c}{\small \underline{Llama}}
    & \multicolumn{2}{c}{\small \underline{PLM}}
    & \multicolumn{1}{c}{\small \underline {QVQ}}
    \\

    &
    \small 7B & \small 32B & \small 72B
    &
    \small 7B & \small 72B
    &
    \small Small & \small Base
    &
    \small V-o1
    & \small 3.2 & \small 4
    &
    \small 3B & \small 8B
    & \small 72B\\

    \hline
    Acc. (\%)
    & 0.1
    & 0
    & 0.01
    & 0.05
    & 0
    & 0.07
    & 0.04
    & 0.1
    & 0.1
    & 0
    & 0
    & 0 
    & 0.03\\
    \bottomrule
\end{tabular}
 \caption{On \datasetComplex, with the complexity ranging from 8 to 16 objects per scene, model performance \textit{severely} degrades. The best performing models reach <1\% accuracy. PLM here stands for PerceptionLM. }
\label{tab:challenge-results}
\end{table}

\paragraph{When objects are similar, it's harder for models.}

Next, we test the effects of the \textit{similarity} of the common objects within a set. If objects in images are similar, it may pose a unique challenge for models. For a given set of common objects, $\mathcal{O}_{common}$, we compute an embedding for each object in the set and take the maximum pairwise similarity as a proxy for object similarity. We use the NV-Embed2 embedding model \citep{lee2024nv}, as it was optimized for embedding similarity. We observe that accuracy generally \textit{decreases} as object similarity \textit{increases}, meaning similarity among objects perhaps makes the task of reasoning about commonality more challenging. We validate this statistically by computing the Pearson correlation between similarity of common objects and accuracy, and find 10 or our 13 tested models have statistically significant, negative correlations of small effect size with $|r| >= 0.3$ (see Appendix, \Cref{tab:obj-sim-correlation} for full results). 

\subsection{How do real and synthetic images compare?}

\begin{figure}[h]
    \centering
    \includegraphics[width=\linewidth]{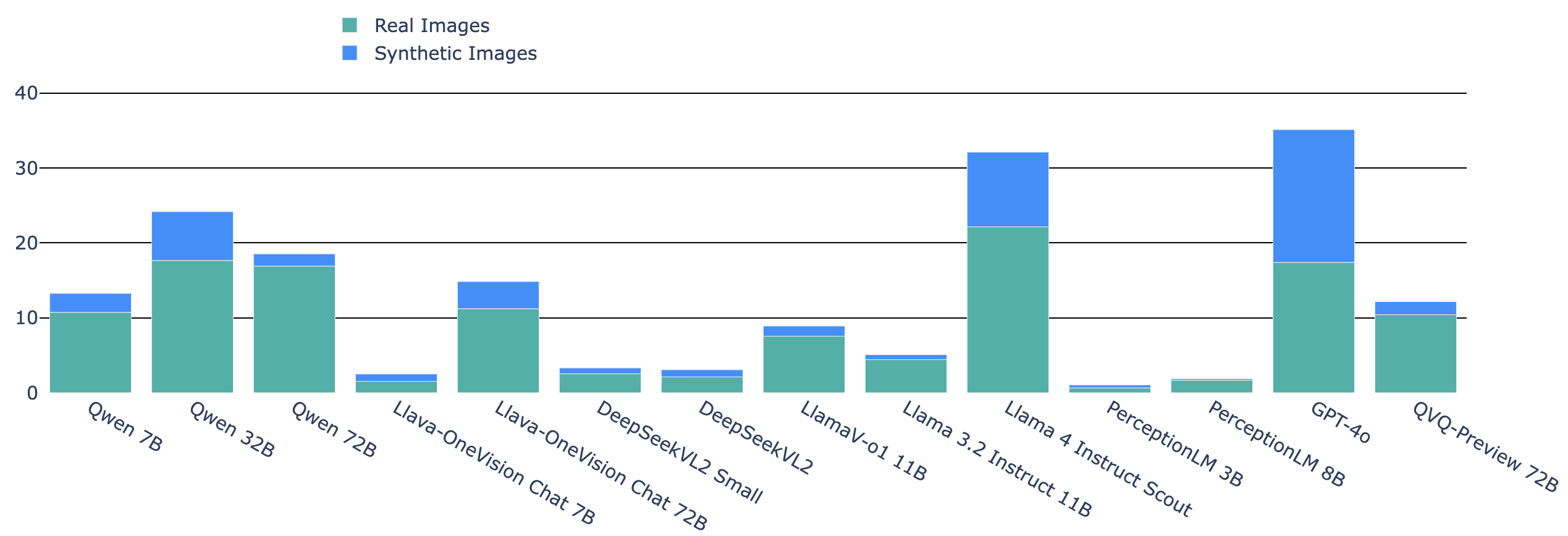}
    \caption{Performance on \dataset\ subsetted according to whether example image pairs are real or synthetic. The height of each bar represents the total accuracy on \dataset: the {\bf\realImageColor green area} of the bar represents the contribution of the real image accuracy, and the {\bf\synImageColor blue portion} of the bar represents the contribution of the the synthetic portion. Models tend to have higher performance on real images (larger green area) than on synthetic ones (smaller blue area). However, the difference in performance on the two subsets decreases as overall accuracy (bar height) decreases, with the DeepSeek-VL2 family, the  PerceptionLM family, Llama 3.2 Instruct 11B, and Llava-OneVision 7B, having only a small difference between the two subsets.}
    \label{fig:real-v-synthetic}
\end{figure}

We compare model performance on the real images versus synthetic images.
To do this, we focus on \dataset\ results, and subset the dataset according to whether the examples were real or synthetic. We find that synthetic images are generally more challenging for models than real images, with less of a gap between the performances on the two data subtypes for models that were less performant overall on \dataset\ (see \Cref{fig:real-v-synthetic} for full results). Though the synthetic images are similar to real images in several respects, having the same scene complexity and using multiple camera orientations per configuration, the synthetic images have the potential to be more diverse in backgrounds and object sizes. This increased difficulty may also indicate a domain shift from models' training data. We used diverse backgrounds (\eg green marble, concrete, aluminum) and relative object sizes that are less common in the real world (\eg a rubber duck being the same size as a remote).
Additionally, because data contamination is difficult to avoid once benchmarks are openly available on the internet, our results show the benefit of leveraging synthetic data without compromising on image difficulty or quality.

\subsection{Models trained on multi-image inputs show improved ability to reason across scenes}

\begin{figure}
    \centering
    \begin{subfigure}[t]{0.32\textwidth}
    \includegraphics[width=\linewidth]{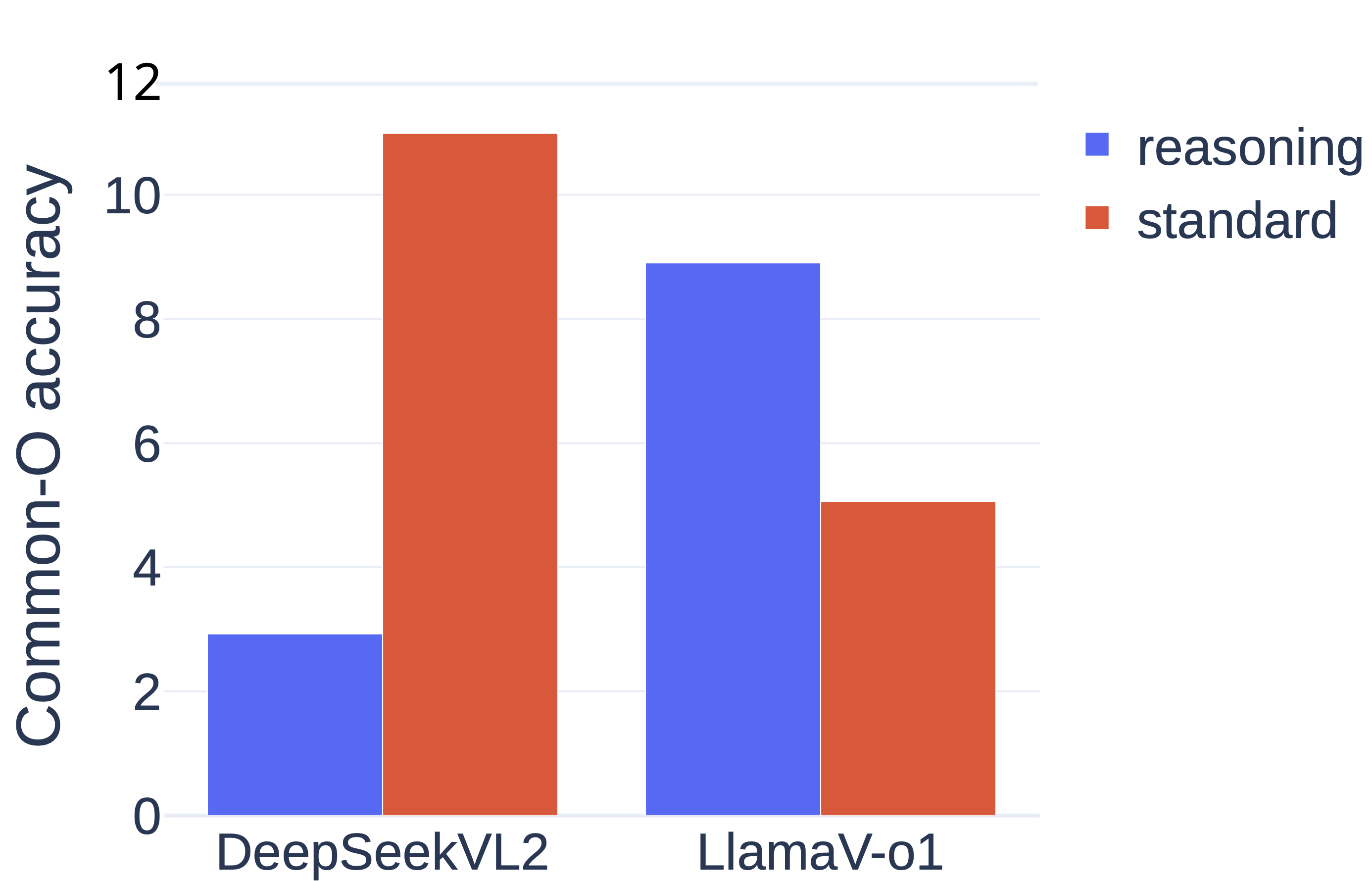}
    \caption{Reasoning Models (CoT)}
    \end{subfigure}
    \begin{subfigure}[t]{0.32\textwidth}
    \includegraphics[width=\linewidth]{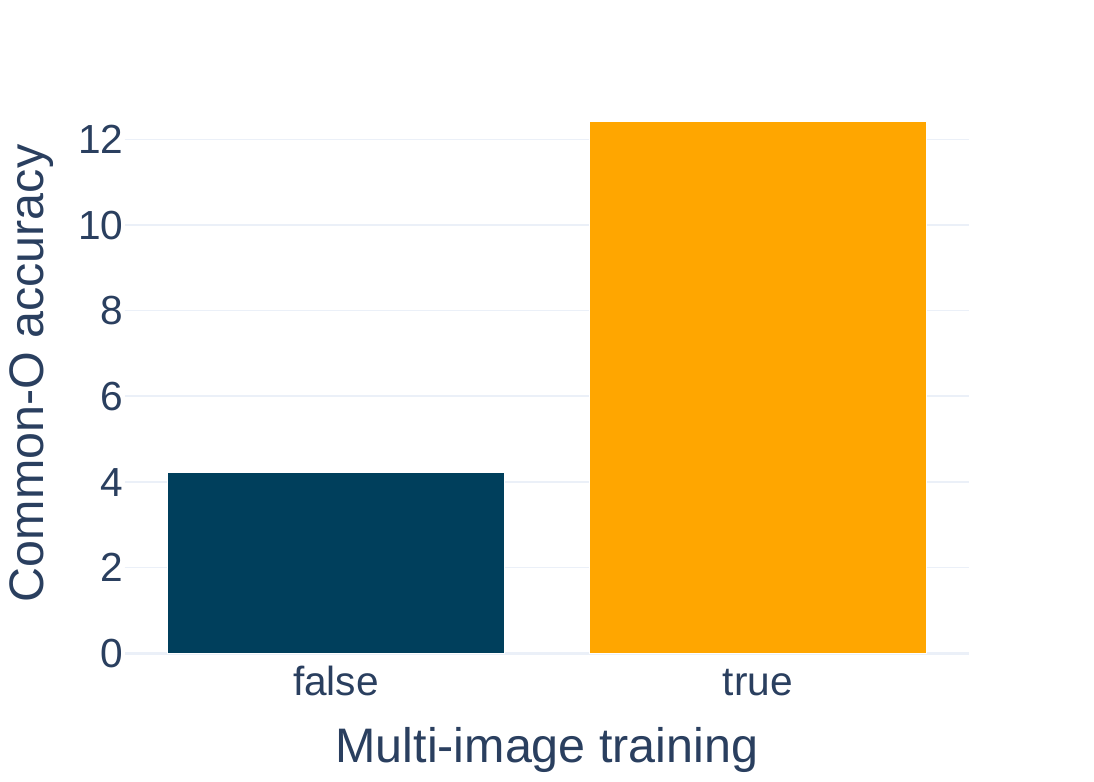}
    \caption{Average Multi-image training}
    \end{subfigure}
    \begin{subfigure}[t]{0.32\textwidth}
    \includegraphics[width=\linewidth]{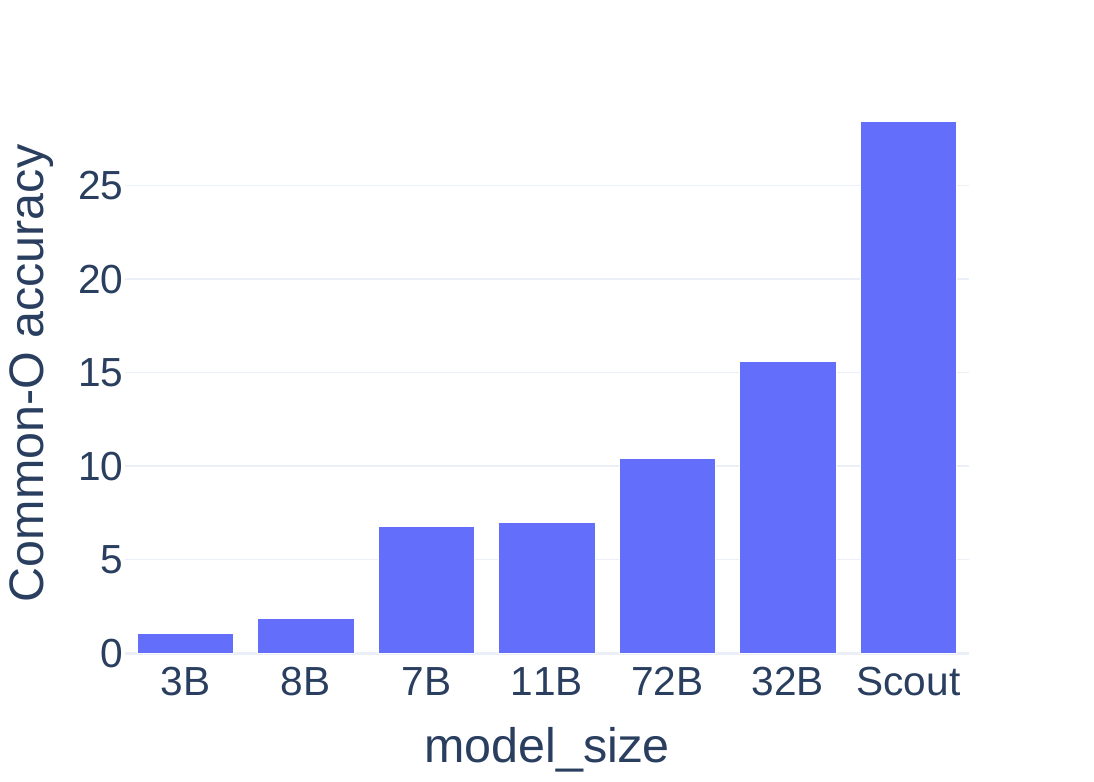}
    \caption{Average model scale}
    \end{subfigure}
    \caption{\textbf{Accuracy on our benchmark in different settings:} In (a), model families differ in whether their reasoning models (with CoT) perform better or works on \dataset. In (b) and (c), we see improved reasoning for models that utilized multi-image training and were larger overall, suggesting using two approaches may enable better performance on \dataset. Note: We average across several models when they have the same size or training-setup.}
    \label{fig:multi-image-training}
\end{figure}

Finally, we explore which levers offer promise for advancing multimodal models' capacity to reason across scenes. We analyze performance based on whether models are explicitly trained on multi-image inputs, with CoT reasoning, and at large scale (many model parameters) in \Cref{fig:multi-image-training}. We find that CoT reasoning, which unlocks ``thinking'' tokens to parse scenes, has a mixed effect on reasoning across scenes, despite boosting single image perception across both model families we studied (78\% for DeepSeek versus 70\% for Qwen and 77\% LlamaV-o1 versus Llama 3.2 Instruct 73\%). This suggests standard reward based reasoning requires further research to enable reasoning across scenes. On the other hand, we see promise in models trained with multi-image inputs have $3\times$ higher accuracy on \dataset\ compared to those trained with single image training. We also, perhaps unsurprisingly found that larger models had stronger performance, which suggests that scaling model size may help boost accuracy.

\section{Discussion}

\paragraph{Limitations.} The real images in our benchmark were all taken by the authors, which understandably may reflect some bias in terms of locations, backgrounds, and objects used. The usage of synthetic image helps include more image diversity. Additionally, multiple choice setups are known to be somewhat brittle \citep{zheng2023large,long2024llms,gupta-etal-2025-changing}---simple changes to prompts and the order of choices can impact performance. An ideal setting would be open-ended generation, where models are able to use describe and reason about objects with their own labels. At present, our datasets only include English text. Multilingual evaluation settings could be interesting future work.

\paragraph{Contributions.}
With multimodal models saturating vision leaderboards focused on perception, we introduce \dataset, a challenge for reasoning across scenes.
We find while perceiving objects in a single image is easy, reasoning across the same scenes is challenging: the best performing model reaches just 35\% accuracy on \dataset\ -- and no model is above 1\% on our challenging subset \datasetComplex. 
We discover models are prone to hallucination when similar objects are present suggesting models may still be relying on object co-occurrence seen during training, as opposed to reasoning in the more flexible way we recognize in humans. To advance the essential skill of reasoning across scenes, new training paradigms that explicitly incorporate multi-image inputs with forms of reasoning going beyond existing reward feedback are called for to overcome the challenge of hallucinations when reasoning across scenes.

\paragraph{Acknowledgments.} We thank FAIR colleagues Olga Golovneva, Kamalika Chaudhuri and Christoph Feichtenhofer for their thoughtful feedback on our paper and for suggesting exciting experiments.

\bibliographystyle{assets/plainnat}
\bibliography{deduplicated}

\clearpage
\newpage
\beginappendix

\section{Image Taking Guidelines}\label{app:takingpictures}
We used the following procedure to guide our creation of images. First, each image taker selected a set of up to $7$ objects and identified a background (e.g. a blanket, counter, or on the floor). Second, they take images iteratively, starting by placing a single object on the background and subsequently adding others (N=1 to N=7). Images were framed with the objects in the center or slightly off center (e.g. in \Cref{fig:challenge-set-example}, the plants in the third set of images from the left has leaves outside of the top part of the frame), with the goal that the majority if not the entirety of the object be contained within the frame. Across scenes, objects are often viewed from different viewpoints (e.g. top-down, versus side-view). Objects also may be partially occluded by other objects in the scene (e.g. in the bottom left image in \Cref{fig:challenge-set-example} the eye-mask is slightly occluded by the pink ball), but occlusions should be minimal with the restriction that all objects be easily human recognizable. For each scene (set of objects against a background), the image-taker would also take images from multiple visual orientations freely (with no restriction on the angle between the camera and the objects, so as to better capture real world diversity). 
Third, the image-taker would repeat against a new background, and add the objects to the scene in a different order and at a different orientation. Throughout this process, image-takers refrained from including any sensitive objects which may have privacy or IP concerns (e.g. humans, animals, brands, logos etc.) in images. Images were taken using smart phone cameras (Google Pixel, iPhone 15 Pro), as smart phones are one of the predominant modes of image creation currently.  

\section{Additional Analysis}

\paragraph{Accuracy for Single Image Perception Versus Multi-Image Reasoning Standard error}
\label{app:standard_errors}

We show in \Cref{tab:single_image_w_std_error} the single image performance with standard error. We report the same performance for the multi-image reasoning task in \Cref{tab:multi_image_w_std_error}. To compute the standard error we run bootstrapping with 1000 iterations on both the single image (baseline) and multi-image settings. Overall, we find a very small standard error.

\begin{table}[]
\begin{tabular}{l|ccc}
\toprule
Model                    & Accuracy & Bootstrap Mean & Standard Error \\
\hline
Qwen 7B                  & 65.16    & 65.21          & 0.02           \\
Qwen 32B                 & 70.34    & 70.35          & 0.02           \\
Qwen 72B                 & 69.18    & 69.17          & 0.02           \\
Llava OneVision Chat 7B  & 78.17    & 69.73          & 0.02           \\
Llava OneVision Chat 72B & 77.7     & 77.17          & 0.02           \\
DeepSeek VL Small        & 78.3     & 78.3           & 0.02           \\
DeepSeek VL              & 69.8     & 69.8           & 0.02           \\
LlamaV-o1 11B            & 77.28    & 77.27          & 0.02           \\
Llama 3.2 Instruct 11B   & 73.42    & 73.42          & 0.02           \\
Llama 4 Instruct Scout   & 80.31    & 80.33          & 0.02           \\
Perception LM 3B         & 79.55    & 79.56          & 0.02           \\
Perception LM 8B         & 84.59    & 84.60          & 0.02           \\
GPT-4o                   & 77.28    & 77.29          & 0.02          \\
\bottomrule
\end{tabular}
\caption{Single image accuracy with standard error using bootstrapping with 1000 iterations.}
\label{tab:single_image_w_std_error}
\end{table}

\begin{table}[]
\begin{tabular}{l|ccc}
\toprule
Model                    & Accuracy & Bootstrap Mean & Standard Error \\
\hline
Qwen 7B                  & 13.26    & 13.27          & 0.01           \\
Qwen 32B                 & 24.16    & 24.14          & 0.01           \\
Qwen 72B                 & 18.53    & 18.55          & 0.01           \\
Llava OneVision Chat 7B  & 2.61     & 2.61           & 0.005          \\
Llava OneVision Chat 72B & 14.84    & 14.85          & 0.01           \\
DeepSeek VL Small        & 3.31     & 3.31           & 0.005          \\
DeepSeek VL              & 3.06     & 3.07           & 0.005          \\
LlamaV-o1 11B            & 8.9      & 8.9            & 0.009          \\
Llama 3.2 Instruct 11B   & 5.07     & 5.08           & 0.006          \\
Llama 4 Instruct Scout   & 35.12    & 35.14          & 0.01           \\
Perception LM 3B         & 1.04     & 1.04           & 0.003          \\
Perception LM 8B         & 1.86     & 1.86           & 0.004          \\
GPT-4o                   & 35.11    & 35.11          & 0.01          \\
\bottomrule
\end{tabular}
\caption{Multi-image reasoning accuracy with standard error using bootstrapping with 1000 iterations.}
\label{tab:multi_image_w_std_error}
\end{table}

\paragraph{Prompt Variants and Temperature}
\label{app:temperature}

We report two additional prompt reformulations (along with temperature ablations) for a total of 3 prompts on Qwen 7B. For a given temperature, we find the overall performance differs by 1.5-2.6\% across prompts suggesting our claims are robust to prompt reformulations. We provide a full table of these results in \Cref{tab:temperature}.

\begin{table}[]
\begin{tabular}{l|ccc}
\toprule
Temperature & Prompt \#1 Acc. & Prompt \#2 Acc. & Prompt \#3 Acc. \\
\hline
0.0         & 13.6            & 11.3            & 12.5            \\
0.2         & 13.2            & 11.0            & 11.3            \\
0.4         & 13.4            & 11.5            & 11.7            \\
0.6         & 13.5            & 11.2            & 11.4            \\
0.8         & 13.1            & 11.6            & 11.9            \\
1.0         & 13.3            & 11.2            & 12.0           \\
\bottomrule
\end{tabular}
\label{tab:temperature}
\caption{We report accuracy across prompt reformulations across six temperatures for Qwen 7B.}
\end{table}

\paragraph{Role of Object Similarity}

In \Cref{tab:obj-sim-correlation}, we show the correlation between accuracy and the average similarity of objects in the scene. We observe a statistically significant negative correlation suggesting as models are more likely to make mistakes when objects are similar.

\begin{table}[h]
\centering
\begin{tabular}{c|c}
    \toprule
    Model & Pearson Correlation \\ \hline
    Qwen 7B & \bf -0.33* \\
    Qwen 32B & \bf -0.38* \\
    Qwen 72B & \bf -0.40* \\
    Llava-OneVision Chat 7B & \bf -0.38* \\
    Llava-OneVision Chat 72B & \bf -0.30* \\
    DeepSeek-VL2 Small & -0.12* \\
    DeepSeek-VL2 & \bf -0.30* \\
    LlamaV-o1 11B & -0.29* \\
    LlamaV 3.2 11B & \bf -0.33* \\
    Llama 4 Instruct Scout & \bf -0.41* \\
    PerceptionLM 3B & -0.10 \\
    PerceptionLM 8B & \bf -0.35* \\ \bottomrule
\end{tabular}
\caption{Correlation between similarity among common objects and accuracy. The negative correlation shows that, the more similar the common objects are lead to lower accuracy.  * indicates statistical significance with correlations of moderate strength or above in bold.
}
\label{tab:obj-sim-correlation}
\end{table}

\paragraph{Additional model examples and mistakes}

In \Cref{app-fig:examples}, we show additional randomly sampled examples from \dataset. In \Cref{app-tab:mistakes}, we show randomly selected mistakes in \dataset\ across all models. The examples show the high degree to which models hallucination objects that are not in the ground truth.

\newcommand{\cell}[1]{\parbox{26em}{\normalsize #1}}
\begin{table}[ht!]
\centering
\resizebox{\textwidth}{!}{
\begin{tabular}{llll}
\toprule
 \it Model & \it Choices (Enumerated by Letter to Model) & \it Ground Truth & \it Prediction \\
 
\midrule
GPT-4o & \cell{[silver grater, No objects are in common, dark chocolate bar wrapped in foil, silver straw, silver whisk, silver knife, tangerine, measuring cup]} & No objects are in common & Measuring cup \\[1.1ex] \\

Llava-OneVision & \cell{[No objects are in common, mallard (fake duck), vase, hammer, calculator, dish, basketball, fakefoodcan]} & B, D, E & D, E, H \\[1.1ex] \\

Qwen & \cell{[dumbbell, mouse, hammer, No objects are in common, football, birdhouse, keyboard, volleyball]} & B, D, G & B \\[1.1ex] \\

PerceptionLM & \cell{[spoon, No objects are in common, orange, glass, keys, lime, fork, popcorn kernel]} & E & A, B, C \\[1.1ex] \\

Qwen & \cell{[dino, candle holder, mallard (fake duck), bowl, volleyball, No objects are in common, shaver, birdhouse]} & D, E, G & D \\[1.1ex] \\

Llama 4 Instruct & \cell{[watermelon, plant, No objects are in common, coffee mug, earbuds, candle snuffer, pen, ball]} & D, E, F, H& D \\[1.1ex] \\

Llama 3.2 Instruct & \cell{[bottle opener, gold jigger, 2-prong serving fork, strainer, paring knife with wooden handle, No objects are in common, gold paring knife, silver jigger]} & A, B, D, G, H & D, G \\[1.1ex] \\

Llama 3.2 Instruct & \cell{[fakefruit, airplane, bowl, No objects are in common, spoon, football, keyboard, mouse]} & C, F, G & D \\[1.1ex] \\

Llama 3.2 Instruct & \cell{[fakefoodcan, vase, volleyball, spoon, kitchenware, No objects are in common, fakefruit, shoes]} & A, B, E, G & A, B \\[1.1ex] \\

Qwen & \cell{[remote, basketball, calculator, No objects are in common, mouse, vase, marker, volleyball]} & C, E, H& C \\[1.1ex] \\

Llama 3.2 Instruct & \cell{[fish bowl, white pill bottle, paint brush, candy cane, No objects are in common, orange pill bottle, lint roller, scissors]} & B, F, H & B, D, F, G, H\\[1.1ex] \\

PerceptionLM & \cell{[No objects are in common, candle, marker, fakefruit, keyboard, mallard (fake duck), bowl, remote]} & B, C, E, G, H& A, B, C\\[1.1ex] \\

GPT-4o & \cell{[cup, mallard (fake duck), vase, No objects are in common, football, candle, volleyball, shoes]} & \parbox{8em}{candle, shoes, vase, volleyball} & shoes, volleyball \\[1.1ex] \\

Llama 4 Instruct & \cell{[spoon, No objects are in common, fakefruit, cast iron, basketball, marker, vase, shoes]} & C, D, G, H & C, D, H \\[1.1ex] \\

Qwen & \cell{[spoon, cast iron, basketball, vase, fakefruit, No objects are in common, marker, shoes]} & D, E, H & E, H \\[1.1ex] \\

Qwen & \cell{[No objects are in common, fakefoodcan, fakefruit, shoes, spoon, vase, volleyball, kitchenware]} & C, F, H & B, C, F \\[1.1ex] \\

Llava-OneVision & \cell{[bowl, keyboard, No objects are in common, marker, remote, fakefruit, candle, mallard (fake duck)]} & A, B, D, G & A, B, C \\[1.1ex] \\

Qwen & \cell{[No objects are in common, pail with handle, burnt orange pot, leaf, black pot, easel, pink pot, watering can]} & B, C, E, G & A \\[1.1ex] \\

DeepSeekVL2 & \cell{[No objects are in common, marker, basketball, calculator, vase, mouse, volleyball, remote]} & B, D, E, F, G, H & A, B, C \\[1.1ex] \\

Llava-OneVision Chat & \cell{[black pot, burnt orange pot, pink pot, pail with handle, No objects are in common, leaf, watering can, easel]} & A, C & A, D, G\\[1.1ex] \\

\bottomrule
\label{app-tab:mistakes}
\end{tabular}}
\caption{Randomly sampled model mistakes in \dataset.}
\end{table}

\newcommand{\subfigthirds}[1]{
    \begin{subfigure}[t]{.30\linewidth}\centering #1 \end{subfigure}
}
\begin{figure}
\centering
\subfigthirds{
    \includegraphics[width=\linewidth]{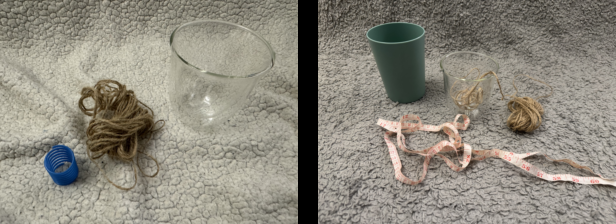}
    \caption{\textit{answer:} glass cup, twine}
}
\subfigthirds{
    \includegraphics[width=0.75\linewidth]{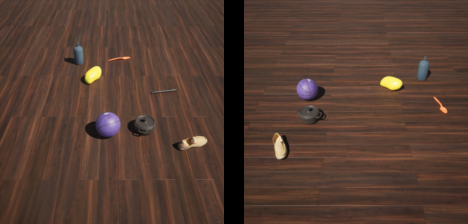}
    \caption{\textit{answer:} basketball, spoon, fakefruit, vase, shoes, cast iron}
}
\subfigthirds{
    \includegraphics[width=\linewidth]{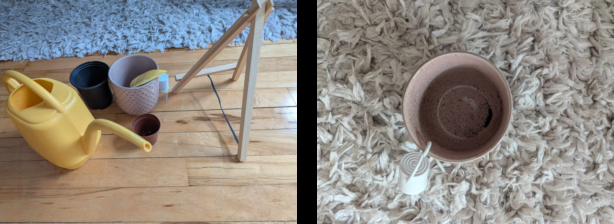}
    \caption{\textit{answer:} pink pot, pail with handle}
}
\subfigthirds{
    \includegraphics[width=\linewidth]{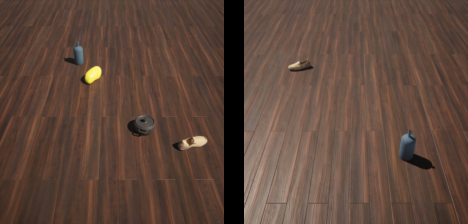}
    \caption{\textit{answer:} vase, shoes}
}
\subfigthirds{
    \includegraphics[width=0.75\linewidth]{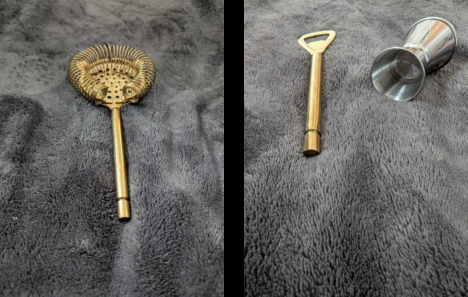}
    \caption{\textit{answer:} No objects are in common}
}
\subfigthirds{
    \includegraphics[width=0.75\linewidth]{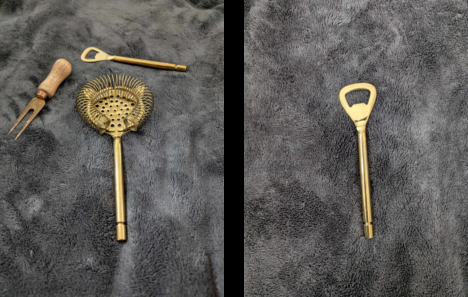}
    \caption{\textit{answer:} bottle opener}
}
\subfigthirds{
    \includegraphics[width=\linewidth]{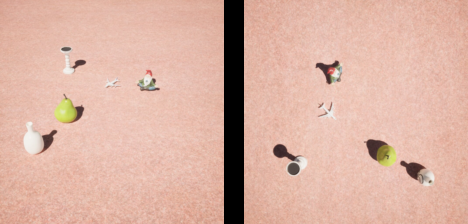}
    \caption{\textit{answer:} fakefruit, airplane, vase, orb, candle holder}
}
\subfigthirds{
    \includegraphics[width=\linewidth]{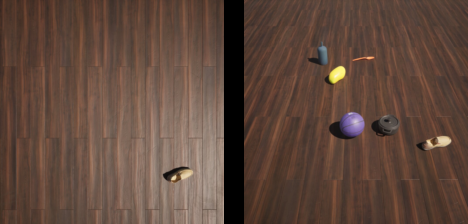}
    \caption{\textit{answer:} shoes}
}
\subfigthirds{
    \includegraphics[width=\linewidth]{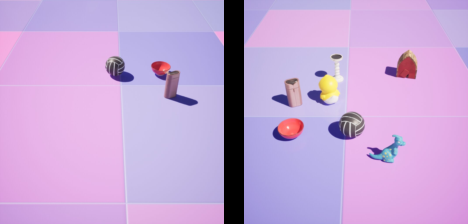}
    \caption{\textit{answer:} shaver, volleyball, bowl}
}
\subfigthirds{
    \includegraphics[width=0.9\linewidth]{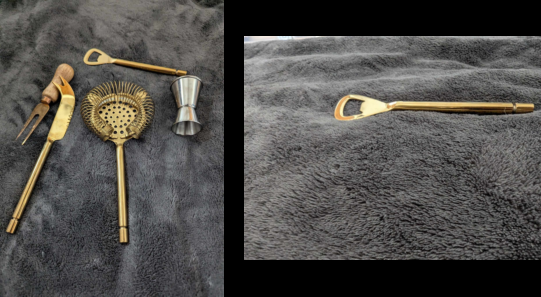}
    \caption{\textit{answer:} bottle opener}
}
\subfigthirds{
    \includegraphics[width=\linewidth]{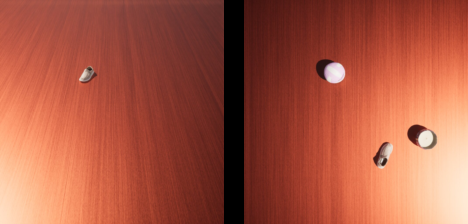}
    \caption{\textit{answer:} shoes}
}
\subfigthirds{
    \includegraphics[width=\linewidth]{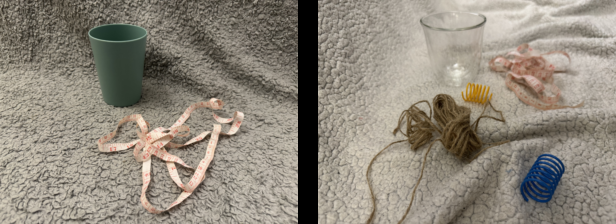}
    \caption{\textit{answer:} measuring tape}
}
\subfigthirds{
    \includegraphics[width=\linewidth]{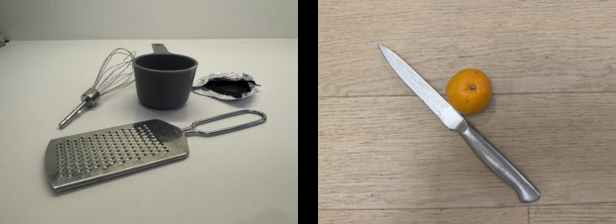}
    \caption{\textit{answer:} No objects are in common}
}
\subfigthirds{
    \includegraphics[width=0.8\linewidth]{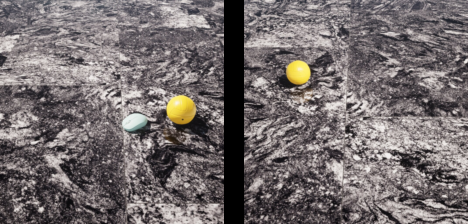}
    \caption{\textit{answer:} volleyball}
}
\subfigthirds{
    \includegraphics[width=0.8\linewidth]{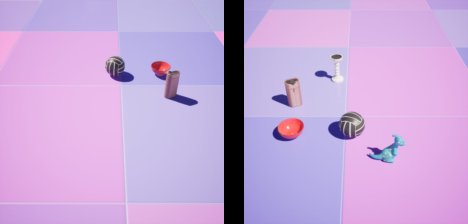}
    \caption{\textit{answer:} bowl, shaver, volleyball}
}
\subfigthirds{
    \includegraphics[width=\linewidth]{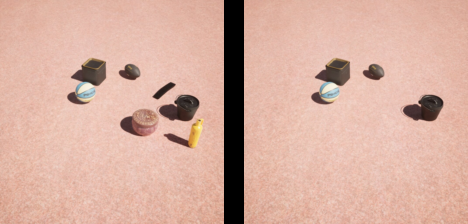}
    \caption{\textit{answer:} kitchenware, basketball, football, cast iron}
}
\subfigthirds{
    \includegraphics[width=\linewidth]{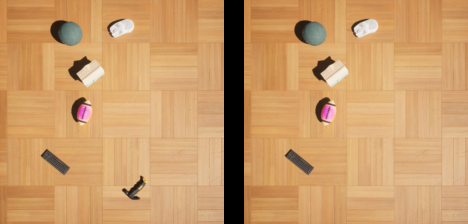}
    \caption{\textit{answer:} football, volleyball, keyboard, birdhouse, mouse}
}
\subfigthirds{
    \includegraphics[width=0.75\linewidth]{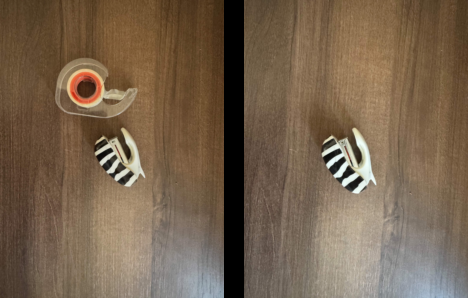}
    \caption{\textit{answer:} Stapler}
}

\caption{Randomly sampled examples from \dataset.}
\label{app-fig:examples}
\end{figure}

\section{Synthetic data}
\label{app:syntheticdata}

The synthetic data was generated using Unreal Engine~\citep{unreal_disclaimer} and assets from Aria Digital Twins Catalog~\citep{Dong_2025_CVPR}. We bought the following asset on fab to get the floor texture with a professional license: \url{https://www.fab.com/listings/66985cc5-13c2-45eb-9b5b-628ef4445a5c}. We randomly placed the assets into one of 16 different positions and apply some slight random rotation over the assets. To ensure that assets are not overlapping with each other, we constrained them to a given maximum size while keeping their aspect ratio. For each scene, we took images coming from 4 different camera positions.

\section{Statistical Analysis of Results}\label{sec:stat-analysis-results}

To get an approximation of the variance, we run bootstrapping with 1000 iterations on both the single image (baseline) and multi-image settings. The results are included below in Table . Overall, we find a very small standard error.

\begin{table}[h]
    \centering\small
    \begin{tabular}{c|ccc|ccc}
    \toprule
        Model &
        \multicolumn{3}{c|}{Multi Image} & 
        \multicolumn{3}{c}{Single Image (Baseline)} \\

        & \it Accuracy & \it  Bootstrap Mean &\it  Std. Err & \it  Accuracy & \it  Bootstrap Mean & \it Std. Err \\\midrule
        
        Qwen 7B & 13.26	& 13.27 & 0.01 & 65.16	& 65.21 & 0.02 \\
        Qwen 32B & 24.16 & 24.14 & 0.01 & 70.34 & 70.35 & 0.02 \\
        Qwen 72B & 18.53 & 18.55 & 0.01 & 69.18 & 69.17 & 0.02 \\
        Llava OneVision Chat 7B & 2.61 & 2.61	& 0.005 & 78.17 & 69.73 & 0.02 \\
        Llava OneVision Chat 72B & 14.84 & 14.85 & 0.01 & 77.7 & 77.17 & 0.02 \\
        DeepSeek VL Small & 3.31 & 3.31 & 0.005 & 78.3 & 78.3 & 0.03 \\
        DeepSeek VL & 3.06 & 3.07 & 0.005 & 69.8 & 69.8 & 0.02 \\
        LlamaV-o1 11B & 8.9 & 8.9 & 0.009 & 77.28 & 77.27 & 0.02 \\
        Llama 3.2 Instruct 11B & 5.07 & 5.08 & 0.006 & 73.42 & 73.42 & 0.02 \\
        Llama 4 Instruct Scout & 35.12 & 35.14 & 0.01 & 80.31 & 80.33 & 0.02 \\
        PerceptionLM 3B & 1.04 & 1.04 & 0.003 & 79.55 & 79.56 & 0.02 \\
        PerceptionLM 8B & 1.86 & 1.986 & 0.004 & 84.59 & 84.60 & 0.02 \\
        GPT-4o & 35.11 & 35.11 & 0.01 & 77.25 & 77.29 & 0.02 \\

        \bottomrule
        
    \end{tabular}
    \caption{Results of running bootstrapping with 1000 iterations. We show the average performance (\texttt{"Accuracy"}) versus the bootstrap mean and standard error on \dataset\ and the single image baseline experiments.}
    \label{tab:placeholder}
\end{table}

\newpage
\clearpage

\section{Dataset Card}\label{sec:dataset-card}




We include a datasheet for \dataset\ below, following the example from \citet{gebru2021datasheets}.

\subsection*{Motivation}
\textit{\lightBlue For what purpose was the dataset created?} The dataset was created the test the reasoning abilities of multimodal LLMs in multi-image, multi-object settings.

\textit{\lightBlue Who created the dataset?}
This dataset was created with contributions from all of the authors on this paper.

\textit{\lightBlue Who funded the dataset creation?} 
This dataset was created with contributions from all of the authors on this paper and funded by Meta.

\textit{\lightBlue Any other comments?} None.

\subsection*{Composition}

\textit{\lightBlue What do the instances that comprise the dataset represent (e.g., documents, photos, people, countries)? Are there multiple types of instances (e.g., movies, users, and ratings; people and interactions between them; nodes and edges)? Please provide a description.}
Each instance is a tuple of 2 images, a set of potential objects that are in both images and a set of the ground-truth, common objects between both images.

\textit{\lightBlue How many instances are there in total (of each type, if appropriate)?}
There are 10586 instances in \dataset\ and 12600 instances in \datasetComplex.

\textit{\lightBlue Does the dataset contain all possible instances or is it a sample (not necessarily random) of instances from a larger set? If the dataset is a sample, then what is the larger set? Is the sample representative of the larger set (e.g., geographic coverage)?}
These were manually created instances, either via the authors taking the images or the authors using a game engine to synthetically create the images. We created a large set of synthetic images ($\approx$400k). For \dataset\ ($N$=3 to $N$=7 objects) and \datasetComplex\  ($N$=3 to $N$=7 objects), we randomly sampled images with the target number of objects.

\textit{\lightBlue Is there a label or target associated with each instance?} The target associated with each instance is the set of objects in common between both images (\eg apple, keys).

\textit{\lightBlue Is any information missing from individual instances?} All of the information is included for every instance.

\textit{\lightBlue Are relationships between individual instances made explicit (e.g., users’ movie ratings, social network links)? If so, please describe how these relationships are made explicit.} Each image in a given contains a specific configuration of objects. This configuration is taken from multiple orientations. These orientations are labeled in the data files. Additionally, each image is contained with multiple instances. The instances in the data file are label with the image filenames so it's clear to see which instances have the same images.

\textit{\lightBlue Are there recommended data splits (e.g., training, development/validation, testing)?}
This is an evaluation-only benchmark; we do not provide any training or validation splits.

\textit{\lightBlue Are there any errors, sources of noise, or redundancies in the dataset?} The instances were manually created. Potential sources of noise may come from ambiguitiy in idenitiying objects, which is captured by our human baseline.

\textit{\lightBlue Is the dataset self-contained, or does it link to or otherwise rely on external resources (e.g., websites, tweets, other datasets)?} The dataset is entirely self-contained.

\textit{\lightBlue Does the dataset contain data that might be considered confidential (e.g., data that is protected by legal privilege or by doctor–patient confidentiality, data that includes the content of individuals’ nonpublic communications)?}
The dataset does not contain any confidential or private information.

\textit{\lightBlue Does the dataset contain data that might be considered sensitive in any way (e.g., data that reveals race or ethnic origins, sexual orientations, religious beliefs, political opinions or union memberships, or locations; financial or health data; biometric or genetic data; forms of government identification, such as social security numbers; criminal history)?}
The dataset does not contain any sensitive information.

\textit{\lightBlue Any other comments?} None.

\subsection*{Collection Process}
\textit{\lightBlue How was the data associated with each instance acquired?} Every real photo was manually taken by one of the authors on this paper specifically for this dataset. Every synthetic photo was generated by the authors using a game engine. We manually wrote the set of objects found in each image.

\textit{\lightBlue What mechanisms or procedures were used to collect the data (e.g., hardware apparatuses or sensors, manual human curation, software, programs, software APIs)?}
We used manual human curation for the real images and the Unreal engine for synthetic images. We validated the images by sampling a subset to hand-annotate.

\textit{\lightBlue If the dataset is a sample from a larger set, what was the sampling strategy (e.g., deterministic, probabilistic with specific sampling probabilities)?}

For the synthetic images, we manually downsampled via random sampling.

\textit{\lightBlue Who was involved in the data collection process (e.g., students, crowdworkers, contractors) and how were they compensated (e.g., how much were crowdworkers paid)?}
The authors performed all components of the data collection.

\textit{\lightBlue Over what timeframe was the data collected?} The data was collected over about 3 months.

\textit{\lightBlue Were any ethical review processes conducted (e.g., by an institutional review board)?}
The data collection went through IRB. We did not include humans in the images.

\textit{\lightBlue Did you collect the data from the individuals in question directly, or obtain it via third parties or other sources (e.g., websites)?}
The data was not collected from external individuals, third parties or web sources. We manually collected all data.

\textit{\lightBlue Were the individuals in question notified about the data collection?}
N/A; see previous question.

\textit{\lightBlue Did the individuals in question consent to the collection and use of their data?}
N/A; see previous question.

\textit{\lightBlue If consent was obtained, were the consenting individuals provided with a mechanism to revoke their consent in the future or for certain uses? If so, please provide a description, as well as a link or other access point to the mechanism (if appropriate).}
N/A.

\textit{\lightBlue Has an analysis of the potential impact of the dataset and its use on data subjects (e.g., a data protection impact analysis) been conducted? If so, please provide a description of this analysis, including the outcomes, as well as a link or other access point to any supporting documentation.}
N/A.

\textit{\lightBlue Any other comments?} None.

\subsection*{Preprocessing/Cleaning/Labeling}

\textit{\lightBlue Was any preprocessing/cleaning/labeling of the data done (e.g., discretization or bucketing, tokenization, part-of-speech tagging, SIFT
feature extraction, removal of instances, processing of missing values)? If so, please provide a description. If not, you may skip the remaining questions in this section}

We manually collected/generated all dataset instances and therefore did not perform any additional data processing beyond image resizing. All images in their original size were saved.

\subsection*{Uses}

\textit{\lightBlue Has the dataset been used for any tasks already?} The dataset has not yet been used in any other tasks.

\textit{\lightBlue Is there a repository that links to any or all papers or systems that use the dataset? If so, please provide a link or other access point.} The dataset is assessible on HuggingFace at \hyperlink{https://huggingface.co/datasets/facebook/Common-O}{this link}.

\textit{\lightBlue What (other) tasks could the dataset be used for?} \dataset\ has been tested for multiple-choice QA with multiple possible answers. The dataset could also be tested in open-ended question answering.

\textit{\lightBlue Is there anything about the composition of the dataset or the way it was collected and preprocessed/cleaned/labeled that might impact future uses?} There is very minimal risk for harm. We did not include any pictures of people, real or generated, and we also excluded any logos. Additionally, this dataset is only for evaluation and therefore will not be used in model training.

\textit{\lightBlue Are there tasks for which the dataset should not be used?} The dataset is exclusively for evaluation and should not be used to train or finetune any models.

\textit{\lightBlue Any other comments?} None.

\subsection*{Distribution}

\textit{\lightBlue Will the dataset be distributed to third parties outside of the entity (e.g., company, institution, organization) on behalf of which the dataset was created? If so, please provide a description.} Yes, the dataset will is publicly available on HuggingFace at \hyperlink{https://huggingface.co/datasets/facebook/Common-O}{this link}.

\textit{\lightBlue How will the dataset will be distributed (e.g., tarball on website, API, GitHub)? Does the dataset have a digital object identifier (DOI)?}
We will host the dataset on HuggingFace. Because this paper is the introduction of the dataset, we will use the paper DOI.

\textit{\lightBlue When will the dataset be distributed?} The dataset is now publicly available and is distributed via HuggingFace.

\textit{\lightBlue Will the dataset be distributed under a copyright or other intellectual property (IP) license, and/or under applicable terms of use (ToU)?} The dataset is being distributed under the non-commercial CC BY-NC 4.0 license.

\textit{\lightBlue Have any third parties imposed IP-based or other restrictions on the data associated with the instances? If so, please describe these restrictions, and provide a link or other access point to, or otherwise reproduce, any relevant licensing terms, as well as any fees associated with these restrictions.} No.

\textit{Do any export controls or other regulatory restrictions apply to the dataset or to individual instances? If so, please describe these restrictions, and provide a link or other access point to, or otherwise reproduce, any supporting documentation.}
No.

\textit{\lightBlue Any other comments?} None.

\subsection*{Maintenance}

\textit{\lightBlue Who will be supporting/hosting/maintaining the dataset?} The paper authors will be maintaining the dataset.

\textit{\lightBlue How can the owner/curator/manager of the dataset be contacted (e.g., email address)?}
Candace Ross and Mark Ibrahim can be contacted through the email addresses provided in the paper.

\textit{\lightBlue Is there an erratum? If so, please provide a link or other access point.}
There is currently not an erratum.

\textit{Will the dataset be updated (e.g., to correct labeling errors, add new instances, delete instances)? If so, please describe how often, by whom, and how updates will be communicated to dataset consumers (e.g., mailing list, GitHub)?}
We will update the dataset for any errors. We will likely communicate this via social media and perhaps a GitHub page.

\textit{\lightBlue If the dataset relates to people, are there applicable limits on the retention of the data associated with the instances (e.g., were the individuals in question told that their data would be retained for a fixed period of time and then deleted)? If so, please describe these limits and
explain how they will be enforced.}
N/A.

\textit{\lightBlue Will older versions of the dataset continue to be supported/hosted/maintained? If so, please describe how. If not, please describe how its obsolescence will be communicated to dataset consumers.
}
N/A

\textit{\lightBlue If others want to extend/augment/build on/contribute to the dataset, is there a mechanism for them to do so? If so, please provide a description.}
We encourage anyone interested in potential augmentations and contributions to contact us using our email addresses, listed above.

\textit{\lightBlue Any other comments?} None.

\end{document}